# Deep Green Function Convolution for Improving Saliency in Convolutional Neural Networks

Dominique Beaini, Sofiane Achiche, Alexandre Duperré, Maxime Raison

**Abstract**
Current saliency methods require to learn large scale regional features using small convolutional kernels, which is not possible with a simple feed-forward network. Some methods solve this problem by using segmentation into superpixels while others downscale the image through the network and rescale it back to its original size. The objective of this paper is to show that saliency convolutional neural networks (CNN) can be improved by using a Green's function convolution (GFC) to extrapolate edges features into salient regions. The GFC acts as a gradient integrator, allowing to produce saliency features by filling thin edges directly inside the CNN. Hence, we propose the gradient integration and sum (GIS) layer that combines the edges features with the saliency features. Using the HED and DSS architecture, we demonstrated that adding a GIS layer near the network's output allows to reduce the sensitivity to the parameter initialization, to reduce the overfitting and to improve the repeatability of the training. By simply adding a GIS layer to the state-of-the-art DSS model, there is an absolute increase of 1.6% for the F-measure on the DUT-OMRON dataset, with only 10ms of additional computation time. The GIS layer further allows the network to perform significantly better in the case of highly noisy images or low-brightness images. In fact, we observed an F-measure improvement of 5.2% when noise was added to the dataset and 2.8% when the brightness was reduced. Since the GIS layer is model agnostic, it can be implemented into different fully convolutional networks. Further, we showed that it outperforms the denseCRF post-processing method and is 40 times faster. A major contribution of the current work is the first implementation of Green's function convolution inside a neural network, which allows the network, via very minor architectural changes and no additional parameters, to operate in the feature domain and in the gradient domain at the same time, thus improving the regional representation via edge extrapolation.

## 1 Introduction

Since the year 2015, the convolutional neural networks (CNN) rose quickly to become the best machine learning technique used to solve the binary problems of computer vision such as edge detection [1, 2], skeleton extraction [3] and salient object detection [4, 5]. In fact, recent algorithms perform near human-level [1].

At first, saliency methods were based on pre-programmed features such as clustering and density [6–8], concavity [9], contrast filtering [10], background detection [11], etc. Although they showed some success with simple images, they did not perform well on more complex dataset images [12]. The method DRFI [13] was the first to use machine learning, but it was soon outpaced by the arrival of CNN-based algorithms with methods such as MDF [14], DCL [5] and DSS [3, 4]. The deeply supervised saliency (DSS) method was successful due to the efficient down-scaling and up-scaling of saliency maps.

An important problem with current salient object detection solutions is that they focus on finding the salient regions with little consideration to the fact that they are often bounded within edges. To overcome this limitation, some methods such as MDF [14] and DCL [5] use a pre-segmentation of the image. Furthermore, most methods fine-tune their results using saliency enhancement methods such as the denseCRF [15] algorithm during the testing phase, which uses segmentation to clean the saliency maps and make it more accurate to the boundaries.

Different methods of enhancing the saliency maps are proposed in the literature. WCtr [11] proposes to improve the saliency maps using background detection. However, BGOF [16] showed that most saliency improvement algorithms based on segmentation and background detection do not work on recent networks since CNN are better at detecting the background and segmentation than traditional methods. In contrast, algorithms such as denseCRF [15] and BGOF [16] optimize the saliency map density via energy minimization. The DeepSets method [17] is very similar since it uses super-pixels to enhance the saliency maps on the boundaries and increase the density. Alternatively, the RACDNN method [18] proposes to use a recurrent attention mechanism to recursively enhance each region of the saliency map. Although RACDNN trains with the network, the attention mechanism is outside it and significantly increases the architectural complexity by adding recurrent layers [18]. As explained in later paragraphs, the proposed method differs from the literature since it adds a layer directly inside the network, thus directly improving the capacity of the network to generate saliency maps.

**The objective of this paper** is to show that a saliency CNN can be improved using a Green's function convolution (GFC), which allows integrating edge-like features into salient features. Hence, we propose the gradient integration and sum (GIS) layer, which integrates the gradient domain features and adds them to the special domain features. By doing so, the GIS layer creates a smooth and continuous region between the high gradient boundaries, thus enhancing the saliency map inside boundaries and reducing it outside the boundaries. Hence, for the proposed method, the network directly trains the parameters used for the saliency improvement, in contrast with other methods which act outside of the main network.

It is to note that the GIS layer aims at improving the saliency in the uniform regions inside boundaries, not at the boundaries themselves, by allowing the network to learn regions via edge filling. This is because the GFC extrapolates the edges into regions of smooth probabilities within the image, as demonstrated by Beaini et al. [19]. This is different then other saliency methods such as denseCRF [15], BGOF [16] and DeepSets [17] which aims at correcting the saliency maps at the boundaries, not inside the boundaries.

The denseCRF [15], BGOF [16], DeepSets [17] and RACDNN [18] are methods that post-process the saliency maps outside of the neural networks, aiming to improve the boundaries, to increase the contrast and the density of the maps. This is in contrast with the proposed method consisting of adding a GIS layer directly inside the network, thus allowing the network to train its inputs. Furthermore, GIS does not aim at improving the density or the boundaries but consists of allowing the network to combine features from the saliency-domain and the gradient-domain.

The GIS is first proposed in this paper, although similar concepts of gradient-domain merging were previously both proposed by Beaini et al [20]. Previous work using Green's function convolution (GFC) for Poisson image editing, contrast enhancements and paint-like effects [20–23]. To our knowledge, they are never implemented inside neural networks. However, the GFC method was demonstrated to solve 100 Laplacians in 1ms using machine learning libraries such as Pytorch and Tensorflow [20].

The idea of merging edges with saliency inside the network comes from the fact that edge detectors are fast to learn by a CNN since they require gradient-like [24] or Gabor-like kernels [25, 26]. Thus we propose to create a network that computes saliency and edge-like features at the same time, then merges them using a GIS layer. Further, to better understand the importance of the edges in the saliency detection, we visualize the inputs and outputs of the GIS layer at different scales. The current paper demonstrates that the proposed GIS layer improves salient object detection for different network architectures, resulting in better accuracy, less overfitting and lower sensitivity to the network initialization.

In our work, we propose to use the GIS layer on the HED [2] and DSS [3, 4] architectures by adding our layers to the end of each side-layer of the original networks, without any other architectural changes. Both HED [2] and DSS [3] are known for their edge detection performance, but only DSS [4] performs well for salient object detection. However, our work shows that the GIS layer improves the HED network by a high margin, thus allowing it to outperform saliency-focused networks. We will refer to the modified models as HED-GIS and DSS-GIS.

Since the proposed GIS layer is directly integrated into the network to improve it's learning capacity, alternative methods that post-process the saliency maps such as denseCRF, BGOF and DeepSets do not compete with GIS, they are complementary. In fact, an original contribution of the current work is the first implementation a GFC-based layer inside any type of CNN, allowing the network to extrapolate edges into smooth regions.

## 2 Methodology

The full implementation of DSS-GIS is done with Python using the TensorFlow library. The current section will explain how the GIS work, what are the HED and DSS architectures and how we modify them to create the proposed HED-GIS and DSS-GIS.

### 2.1 Gradient integration and sum (GIS)

The GIS method is first proposed in the current work but is inspired by the field of *gradient-domain image editing*, which mainly focuses on applying editing filters to images [27, 28]. We mainly base our work on the GFC method proposed by Beaini et al. [20] which allows integrating any vector field with minimal error and low computation time (100 images in 1ms). For the current paper, we are interested in the ability of the GIS to combine edge-like features (object boundaries) with region-like features (saliency maps). Those features cannot be combined by standard operations such as additions, multiplications or small-kernel convolutions since most edge pixels do not intersect saliency pixels. Hence, before summing the features, an integration step is required to transform the edge-like features into region-like features. The following subsections will explain how the GIS integrates the gradient and merges it with the standard features.

### 2.1.1 Green's function convolution (GFC)

The current subsection focuses on the gradient-integration step and is based on work by Beaini et al. [20].
Let us denote $E$ as a vector field of features made of the horizontal $E_x$ and vertical $E_y$ components. The vector field $E$ cannot be integrated directly since it is not necessarily a conservative field, meaning that it does not have a solution.

Hence, we use Green's function based solver proposed by Beaini et al. [20]. We first need to compute the Laplacian $L_p$, then to solve it using a Green's function convolution (GFC), as described in this section.

The Laplacian $L_p$ is computed using equation (1), where $E_{x,y}$ are the $x$ and $y$ components of the field $E$ and $K_{E \to L}$ is the convolutional kernel that represents this operation.

$$L_p = \frac{\partial E_x}{\partial x} + \frac{\partial E_y}{\partial y} = K_{E \to L} * E \qquad (1)$$

Now that the Laplacian is computed, we need to compute the Green's function that solves it. The Green's function is defined as a function that solves a given differential equation with a convolution [29]. In our case, the differential equation is the numerical Laplacian given by the convolution in equation (2). In this equation, $I$ is any image, $L_I$ is its Laplacian and $K_{\nabla^2}$ is the Laplacian kernel.

$$L_I = I * K_{\nabla^2} \; , K_{\nabla^2} = \begin{bmatrix} 0 & -1 & 0 \\ -1 & 4 & -1 \\ 0 & -1 & 0 \end{bmatrix} \qquad (2)$$

If we denote $V_{mono}$ as being the numerical Green's function that solves the Laplacian, then equation (3) shows that the convolution $K_{\nabla^2} * V_{mono}$ act as an identity. Since the convolution identity is the Dirac's delta $\delta$ [29], then equation (4) represents this relation.

$$I = I * K_{\nabla^2} * V_{mono} \qquad (3)$$

$$K_{\nabla^2} * V_{mono} = \delta \qquad (4)$$

Then we define the convolution theorem [29] in equation (5) where $A, B$ are any function, $\mathcal{F}$ is the Fourier transform, $\mathcal{F}^{-1}$ is the inverse Fourier transform and $\circ$ is the element-wise product.

Using equation (5) it becomes possible to solve equation (4) for $\mathcal{F}(V_{mono})$, as given by equation (6). The notation $V_{mono}^{\mathcal{F}}$ represents the Green's function in the Fourier domain.

$$A * B = \mathcal{F}^{-1}\big(\mathcal{F}(A) \circ \mathcal{F}(B)\big) \qquad (5)$$

$$V_{mono}^{\mathcal{F}} = \mathcal{F}(V_{mono}) = \frac{\mathcal{F}(\delta)}{\mathcal{F}(K_{\nabla^2})} \qquad (6)$$

Equation (6) gives a solution for the Green's function $V_{mono}$. However, to be applied on an image as given by equation (3), $V_{mono}$ must be the same size as the image.

Hence, we define $\check{\delta}$ as the padded numerical Dirac's delta and $\check{K}_{\nabla^2}$ as the padded numerical Laplacian kernel in equation (7), where $L_I$ is the Laplacian to solve [20]. Then, the Green function in the Fourier domain $\check{V}_{mono}^{\mathcal{F}}$ is given by (8) [20].

$$\check{K}_{\nabla^2} \equiv \begin{bmatrix} 0 & -1 & 0 & \cdots & 0 \\ -1 & 4 & -1 & & \\ 0 & -1 & 0 & & \\ \vdots & & & \ddots & \\ 0 & & & & 0 \end{bmatrix}_{\underbrace{}_{size(L_I)}} \check{\delta} \equiv \begin{bmatrix} 0 & 0 & 0 & \cdots & 0 \\ 0 & 1 & 0 & & \\ 0 & 0 & 0 & & \\ \vdots & & & \ddots & \\ 0 & & & & 0 \end{bmatrix}_{\underbrace{}_{size(L_I)}} \qquad (7)$$

$$\check{V}_{mono}^{\mathcal{F}} = \frac{\mathcal{F}(\check{\delta})}{\mathcal{F}(\check{K}_{\nabla^2})} \qquad (8)$$

In equation (8), $\check{V}_{mono}^{\mathcal{F}}$ is the Green's function that allows to solve any Laplacian by a convolution [20, 29]. The convolution is computed using the Fourier domain as defined in equation (9) since Fast Fourier Transforms (FFT) are faster for large convolutions and are implemented on a graphical processing unit (GPU) in multiple machine vision libraries such as OpenCV [30], MATLAB® [31] and Tensorflow. In equation (9), $\mathcal{R}$ is the real part of a complex number, $c$ is an integration constant and $I_R$ is the resulting image. In practice, a 4-pixels padding of value 0 is added all around $L_p$ to avoid discontinuities in the numerical Laplacian [20]. The constant $c$ is equal to the values in the padded part of $I_R$.

$$I_R = \mathcal{R}\left(\mathcal{F}^{-1}\big(\mathcal{F}(L_p) \circ \check{V}_{mono}^{\mathcal{F}}\big)\right) - c \qquad (9)$$

More details about the mathematical foundation of the Laplacian solver, as well as empirical demonstrations and pseudo-codes are provided in a previous work by Beaini et al. [20].

### 2.1.2 Overview of the GIS layer

To better understand the GIS layer, a graph is provided in **Fig. 1**. We observe that GIS has $n$ output channels for $3n$ input channels. The 3 input groups are $S$, $E_x$ and $E_y$. $S$ is considered in the spatial domain and is simply summed to the output. $E$ is considered in the gradient domain and is integrated using GFC before being summed to $S$.

Note that a weighted sum is not required since the inputs are expected to be weighted by the CNN.

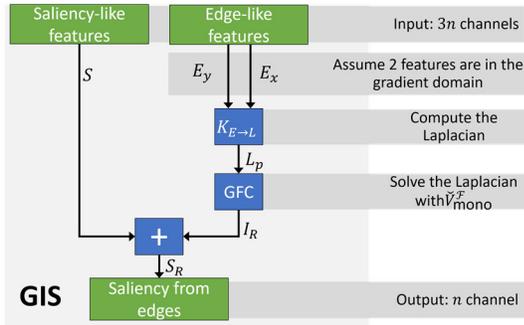

**Fig. 1** Graph summary of the gradient integration and sum (GIS) layer, which outputs $n$ channels from $3n$ inputs.

## 2.2 Implementing the models with the GIS layer

The proposed GIS layer can only be implemented on fully convolutional networks since they require that the network is able to output both saliency and edges at the same time. Hence, we use the DSS model, which to the best of our best knowledge, is one of the most successful saliency model [3, 4]. We also use the HED model to demonstrate that our approach is generalizable to more networks.

It is to note that GIS implements an integration of edge-like features from the gradient domain. However, the network is never forced, via an intermediate loss, to learn the gradient of the saliency map. The gradient of the saliency is naturally learned by the fact that the GFC gradient integrator is used, without requiring a saliency ground-truth. Hence, the edge-like features from **Fig. 1** ($E_x$ and $E_y$) are not necessarily the gradient of the saliency, although they are expected to be. In addition, the saliency-like features used by GIS in **Fig. 1** ($S$) is not forced to be similar to the saliency. It is simply expected to be similar to the saliency due to the merging operation. Hence, $S$ cannot be used directly as an output of the network.

### 2.2.1 The HED and DSS models

The HED and DSS models are an architecture nested on top of a classification network, with deep side layers connected before every pooling [4]. They are presented in **Fig. 2**, with the classification network being the pre-trained VGGnet-16 [32] presented in gray in **Fig. 2**. They have a total of 6 side outputs with 3 layers each, with the first 2 layers being followed by a ReLU operation [4]. The side layers are presented in blue in **Fig. 2** with the parameters defined in a later section in **Table 1**. The only difference between the standard model and our model are that the 3$^{rd}$ layer of each side layer has only 1 output for HED/DSS instead of 3 outputs for HED/DSS-GIS. The weights of the side layers are initialized as a normalized uniform random distribution.

Another innovative concept introduced with DSS but lacking from HED is the short connections between the side outputs. These short connections take the final output of each side layer numbered $n$ and concatenate it with the 3$^{rd}$ layer output of each side layer numbered $m$, where $m < n$. This means that the results from the deeper layers, which are better at finding salient regions, are scaled up and sent to the shallower layers which are better at finding the fine details and edges [4]. These short connections are represented by the blue lines on **Fig. 2**.

This figure highlights the main difference between GIS and state-of-the-art saliency improvement methods such as denseCRF[8], BGOF [16] and deepSets [17]. The proposed GIS layer is implemented directly inside the network, since it is meant to increase the learning capacity of the network. The alternate methods are used as a means to post-process the saliency map, and can still be used alongside GIS.

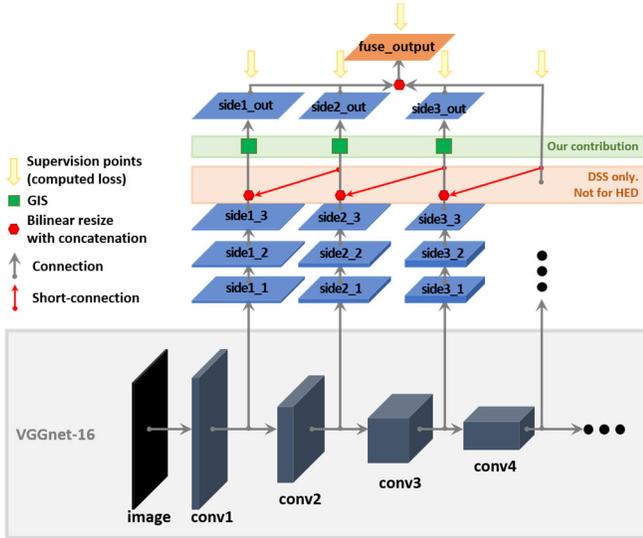
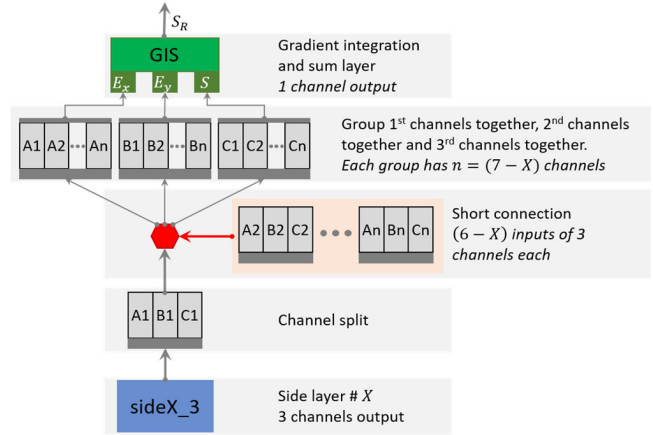

**Fig. 2** The HED [2] and DSS [3, 4] architectures nested upon the pre-trained VGGnet-16 [32] network, with a total of 6 side layers. The red arrows are the short connections implemented by the DSS model [3, 4]. Our contribution is the GIS at the end of each side-layer, which requires the layer sideX_3 to output 3 channels instead of 1.

**Fig. 3** Closer view on the integration of the GIS layer inside the DSS architecture. For the HED architecture, the GIS layer is placed directly after the channel split since there are no short connections.

### 2.2.2 Adding the GIS layer

As stated previously, the main change to the HED/DSS model is the added GIS which allows merging the salient object detection with the salient edge detection. Hence, we use the same side-layers as the HED/DSS method, except that the sideX_3 layers have 3 outputs instead of 1 output. These 3 outputs are then used as inputs to the GIS layer and are also used for the short connection.

A closer view on the integration of the GIS layer within the network is presented in **Fig. 3**, where we observe that each side layer sideX_3 produces 3 outputs, which are then split into the $S$, $E_x$ and $E_y$ inputs of the GIS layer.

All the parameters of the DSS-GIS are summarized in **Table 1** and the architecture is summarized in **Fig. 2**.

For the maximum performance of the DSS-GIS, the GIS layers are expected to have one saliency-like input in the spatial domain and 2 inputs in the gradient domain, since it is how the GIS layer was designed. This is indeed what is observed in **Fig. 4** where $S$ distinguishes the regions and $E_{x,y}$ highlight the edges of the people sitting in the grass.

**Table 1** Side layer information of HED and DSS architectures given by $(n, k \times k)$, where $n$ is the number of output channels and $k \times k$ is the size of the kernel. "Layer" is the name of the layer from the VGGnet-16 whose output is connected to a side layer. "1", "2" and "3" represent the 3 layers for each side output. "1" and "2" are followed by a ReLU operation. If a GIS layer is added, $n_o = 3$, otherwise $n_o = 1$.

| No. | VGG layer | sideX_1 | sideX_2 | sideX_3 |
| --- | --- | --- | --- | --- |
| 1 | conv1_2 | $128, 3 \times 3$ | $128, 3 \times 3$ | $n_o, 1 \times 1$ |
| 2 | conv2_2 | $128, 3 \times 3$ | $128, 3 \times 3$ | $n_o, 1 \times 1$ |
| 3 | conv3_3 | $256, 5 \times 5$ | $256, 5 \times 5$ | $n_o, 1 \times 1$ |
| 4 | conv4_3 | $256, 5 \times 5$ | $256, 5 \times 5$ | $n_o, 1 \times 1$ |
| 5 | conv5_3 | $512, 5 \times 5$ | $512, 5 \times 5$ | $n_o, 1 \times 1$ |
| 6 | pool5 | $512, 7 \times 7$ | $512, 7 \times 7$ | $n_o, 1 \times 1$ |

In **Fig. 4**, We observe that $S_R$ is mainly driven by the $S$ input for the side layer #5 where the resolution is low, but it is mainly driven by the integration over $E_{x,y}$ for the side layers #3, 4 where the resolution is high. This is because the convolutional kernels are too small to detect regions for the high resolution layers. However, they are able to detect edges, which can be integrated into regions via the GIS layer.

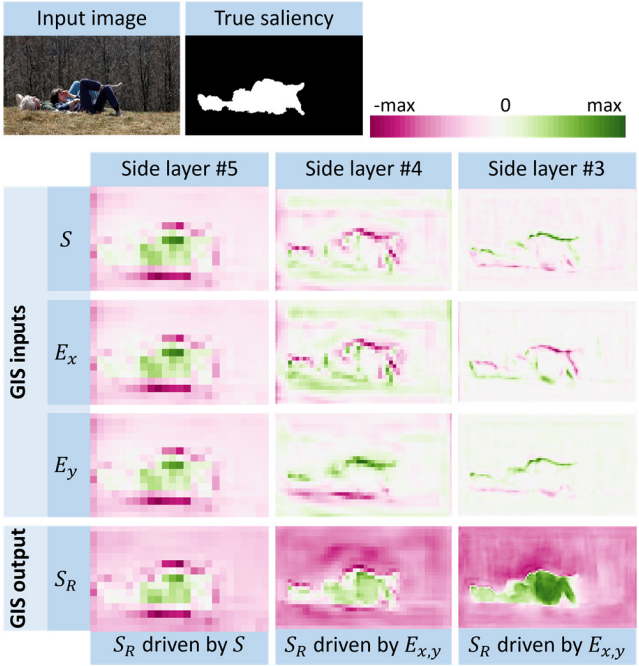

**Fig. 4** Example of the inputs of the GIS layer coming from a fully trained HED-GIS network. $S$ is expected to be in the saliency domain; $E_x, E_y$ are expected to represent the 2 components of the gradient domain.

Other saliency improvement methods do not change the behavior of the network since most of them are about post-processing the saliency map. In contrast, one advantage of the proposed GIS layer is observed in **Fig. 4**, where the gradient integration allows the high resolution layers of the network to learn to create a region from boundaries. Hence the output of the GIS layer ($S_R$) produces a far better saliency map than the regular saliency map ($S$) at medium and high resolution.

In summary, the GIS layers act in a similar way to an activation function at the deepest side-layers of the network since they perform a pre-defined operation on the input channels. However, as shown in **Fig. 1**, GIS outputs a third of its input channels, and it performs a gradient integration operation with a summation. Hence, the GIS layer does not use any weight or intermediate loss.

### 2.2.3 Training procedure

An important modification from the DSS model is that the original code is in Caffe [4] but we recoded the entire architecture in Tensorflow to make use of its multi-platform capabilities, the integrated Fourier transforms, the improved convergence algorithms and the real-time validation curves.

For the parameters, the DSS model [4] proposes to use 10 images per mini-batch. It also uses a standard gradient descent with a learning rate of $10^{-8}$ for 20k iterations and $10^{-9}$ for an additional 4k iterations.

In contrast, we changed those parameters to 8 images per mini-batch, with an Adam optimizer [33] and a learning rate of $4 \cdot 10^{-5}$ for 30k iterations. We use an early stopping method to save the model with the highest F-measure (defined in section 3) on the validation set, then fine-tune the new model for 2000 iterations using a learning rate of $4 \cdot 10^{-6}$.

Those changes are made since our loss is computed as the average loss over the pixels instead of the sum, and because the Adam optimizer removes the need for changing the learning rate [33].

We also use the MSRA10K [34] for training purposes, which is an extension of the previous MSRA-B [13] used for DSS [4]. The MSRA10K is randomly split into 7000 training images, 2000 validation images and 1000 test images. Furthermore, the training images are duplicated using horizontal reflection leading to 14000 training images, as proposed by Hou et al. [4].

Finally, another change that is made to the model is that we use a zero-padding all-around the training images until they reach a resolution of $416 \times 416$. Since every image of the MSRA10K dataset has a maximum resolution of $300 \times 400$, this operation does not resize or crop the images. Furthermore, the computation of the loss, as well as the other measures presented in section 3, ignore all the padded pixels.

## 3 Evaluation datasets and metrics

To evaluate our proposed DSS-GIS algorithm, we need to use standard datasets and metrics. For the datasets, we use the MSRA10K [34] for training since it has the largest number of images (10,000). It is also the most used for training purposes [4]. We randomly split the MSRA10K into 7000 images for training, 2000 images for validation and 1000 images for testing.

For testing purposes, we use the following 3 datasets: ECSSD with 1000 images [35, 36], PASCAL-S [37] with 850 more complex images and DUT-OMRON with the most complex 5168 images [12].

Recently, a new dataset called DUTS [38, 39] was released, containing 10,553 training images and 5,019 test images. This dataset allowed recent methods such as AFNet [40] to have higher performances for salient object detection. We also used this dataset for training and testing purposes for comparison with AFNet.

For the purpose of comparing the performances to other techniques, the parameters that are evaluated are the precision $P$, the recall or true positives $R$ and the false positives $^1R$ [6, 41]. Those parameters are evaluated for 256 levels of thresholds on the saliency map $S$, which allows to plot the precision-recall $PR$ curve. At each threshold level, a binary mask $M$ is generated and compared to the binary ground-truth $G$. From the $PR$ curve, one can evaluate the average $\overline{PR}$, the F-measure $F_m$ and the maximum precision $P_{\max}$. All those parameters are defined in equations (10)-(14), where $\beta = 0.3$ is a constant that allows to add more weight to the precision, $!$ is the logical NOT operator, $\cap$ is the logical AND operator and $\sum$ is the sum over every pixel [6, 41].

$$P = \frac{\sum M \cap G}{\sum M} \quad (10)$$

$$R = \frac{\sum M \cap G}{\sum G}, \quad {}^!R = \frac{\sum M \cap {}^!G}{\sum {}^!G} \quad (11)$$

$$P_{max} = max(P) \quad (12)$$

$$F_m = max\left(\frac{(1+\beta^2)(P\ R)}{\beta^2\ P\ +\ R}\right) \quad (13)$$

$$\overline{PR} = \int P\ dR \quad (14)$$

Other important information is the area under the curve (AUC) of the true-false-positive curve (15), the mean absolute error (MAE) (16), the root mean square error (RMSE) (17) and the cross-entropy (CE) (18) [25]. In those equations, $S$ is the saliency map normalized to $[0, 1]$, $N$ the total number of pixels and $G$ is the ground-truth with binary value 0 or 1.

$$AUC = \int R\ d^!R \quad (15)$$

$$MAE = \frac{1}{N}\sum |S - G| \quad (16)$$

$$RMSE = \sqrt{\frac{1}{N}\sum |S - G|^2} \quad (17)$$

$$CE = \frac{-1}{N}\sum G\ log\ S + (1-G)\ log(1-S) \quad (18)$$

From all those parameters, the most used in the literature are the precision-recall $PR$ curve, the F-measure $F_m$ and the area under the curve $AUC$ [5, 7, 8, 14, 41, 42]. These metrics are used since they represent better the effects of thresholds on the saliency maps and they cannot be improved with simple methods such as contrast enhancement. Hence, those parameters will be used to be compared with other methods from the literature. Additionally, we use the mean precision-recall $\overline{PR}$ and the mean absolute error MAE to show that our approach improves many different measures.

It is worth noting that the cross-entropy $CE$ is used as the loss function for the training of the DSS-GIS model.

## 4 Results

This section presents the saliency map results and a comparison of the validation curves. The results show that DSS-GIS has better saliency maps then DSS, trains faster, is less prone to overfit and has higher accuracy.

### 4.1 Saliency maps

The improved performance of our HED/DSS-GIS approach compared to the standard HED/DSS can be observed on some test images from the ECSSD, PASCAL-S and DUT-OMRON images in **Fig. 5**. In this image, we see that the GIS layer improves the results when there is a bright contrast, a complex background, a camouflaged animal or small objects.

The improvements are very notable when looking at how much the HED-GIS outperforms HED by providing smoother and more accurate saliency maps in all examples. This is because the HED model lacks the upward scale introduced by the DSS short connections required for a good saliency prediction. However, since HED is a good edge predictor, the HED-GIS was still able to produce accurate saliency maps by extrapolating the edge-like features into the image space.

The improvements of DSS-GIS over DSS are subtle. We notice the improvement only for the most difficult examples presented in **Fig. 5**. DSS-GIS is better at discriminating between the salient object and a given background, and better at finding a camouflaged animal or small objects. The reason is that, in these cases, relying on the globally strong edges is more important than relying on the contrast and texture, since the background has a similar complexity as the foreground.

Furthermore, for the *incomplete saliency* category, the DSS-GIS allows finding more salient regions than DSS. The reason is that it fills up the missing saliency regions by ensuring that all the areas inside edges are included in the saliency map, as explained in previous work [19].

However, there are some failures cases of DSS-GIS presented in **Fig. 5**. Those include images where the background has very strong defined edges, but the foreground does not. Also, DSS seems to perform better on transparent objects, since those objects are better detected by their glare than by their edges. It is to note that those failure cases do not apply for HED-GIS, which consistently outperforms HED.

## 4.2 Validation curves

The first difference that we notice when training the proposed DSS-GIS model is that its validation curves are far more similar, faster to train and less prone to overfitting than the DSS model. This can be observed in **Fig. 7** with 6 different training curves of the DSS in orange and 6 curves of the DSS-GIS in blue. Note that these curves are for a learning rate of $10^{-5}$ and different parameter initialization.

The DSS curves are **our** implementation of the model, meaning that it uses exactly the same code as the DSS-GIS, but without the GIS layer.

These curves are generated by computing the loss, the F-measure, the MAE and the RMSE at every 50 iterations, and by randomly selecting 200 images from the validation set. To speed up the computation, the F-measure uses 51 different thresholds instead of the standard 256 levels. Then, an exponential smoothing with a factor 0.9 is applied to all the curves to reduce the noise.

On **Fig. 7**, we see that the DSS-GIS reaches a higher performance for the 4 different measures. Also, the DSS model has a big disparity between different validation curves, meaning that it is more sensible to the random initialization of the side layers and the different initialization algorithms. In fact, DSS does not always converge to its maximal performance if the initialization is not optimal. Finally, we can see that the CE loss of the DSS diverges at around 10k iterations but remains almost constant for DSS-GIS.

All those differences show that the DSS-GIS is easier and faster to train, less prone to overfit and leads to better results than the DSS model. Furthermore, the training is more robust to the random parameter initialization, leading to more similar training curves across different trainings. Since the only difference between the 2 models is a fast to compute GIS layer at each side output, we deduce that our proposed DSS-GIS outperforms the DSS model.

After optimizing the initialization parameters and the learning rate via cross-validation, we obtained the curves presented in **Fig. 6**. In this figure, we observe that the optimal DSS converges as fast as DSS-GIS and has less overfit than the models in **Fig. 7**. However, it still converges to a lower validation performance. We also observe that the HED-GIS model strongly outperforms the standard HED model.

In addition, the GIS layer seems to reduce the noise of the validation curves, meaning that the network converges more easily to the optimal solution. Hence, we conclude that the GIS layer allows to significantly improve the training process of the networks.

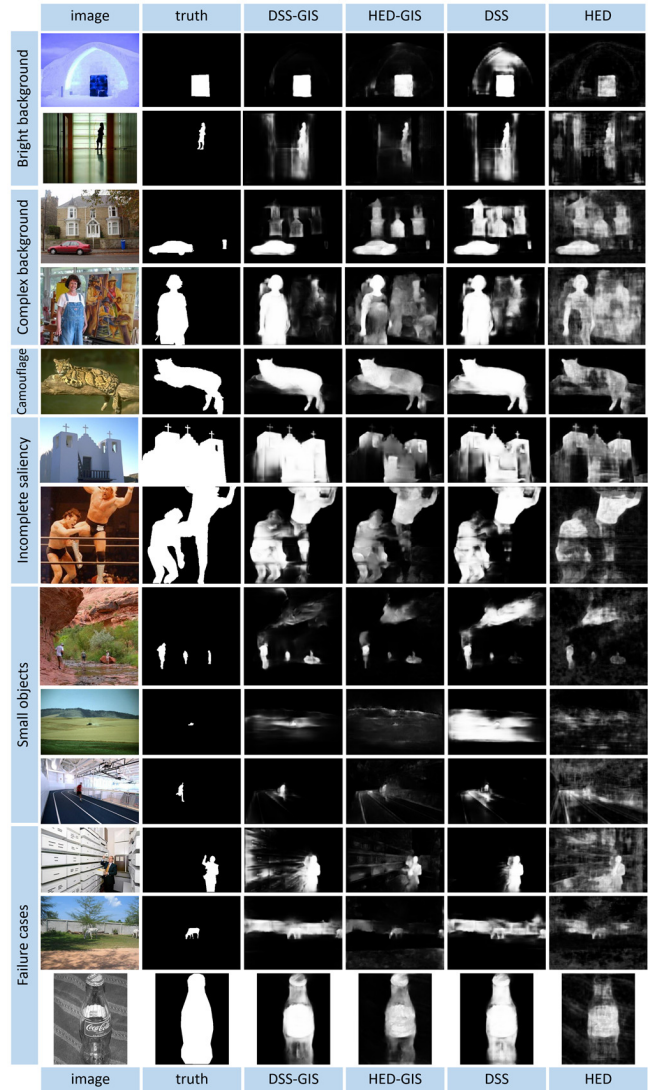

**Fig. 5** Test results comparison between the DSS and HED methods with and without the proposed GIS layer. The training set is MSRA10K.

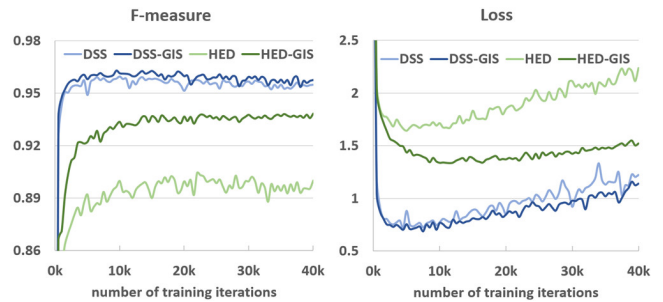

**Fig. 6** Comparison of the validation curves of different models for the optimally found parameters. The validation performance is computed every 500 iterations on the full validation set. The training set is MSRA10K.

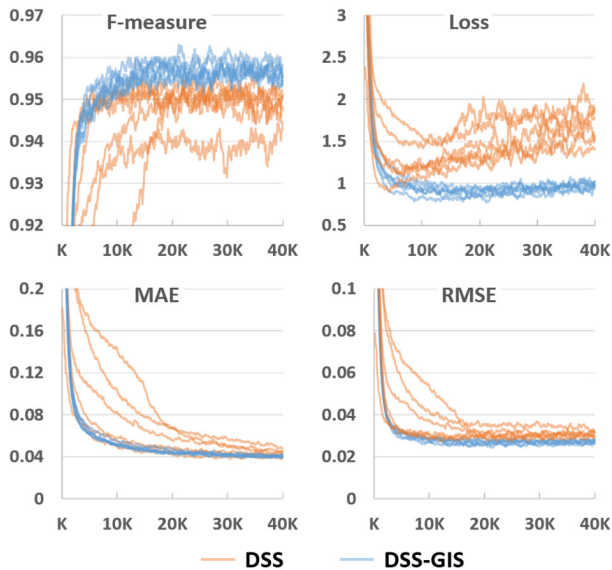

**Fig. 7** Comparison of 6 validation curves in orange of our implementation of the DSS model, and 6 validation curves in blue of our DSS-GIS model. The curves are smoothed using exponential smoothing with a factor 0.9, and the x-axis represents the number of iterations. The training set is MSRA10K.

## 5 Literature comparison and discussion

This section will compare our proposed HED/DSS-GIS models to multiple models in the literature using the metrics specified in section 3. It will show that the DSS-GIS model outperforms the other model on every dataset.

### 5.1 Training the DSS, DSS-GIS

For a fair comparison between the standard method and the same method with the GIS layer, we use exactly the same training procedure as the one defined previously in section 2.2.3. Furthermore, the random seed is the same for all models, meaning that the network initialization is identical across models and the training-validation-testing split is also identical.

### 5.2 Testing improvement of the GIS layer

In this section, we compare the results given by the DSS and HED models with and without the proposed GIS layer.

For the DSS method, our implementation performs better than the original implementation in their paper [42]. For a fair comparison, we thus use our implementation of DSS for all reported test results.

The compared results are shown in **Table 2** for HED and **Table 3** for DSS, where the * symbol means that a denseCRF layer is added. We observe that the proposed DSS-GIS is always better than DSS, and that the proposed HED-GIS is always better than HED.

**Table 2** shows that GIS always outperforms denseCRF for all metrics for the HED model. **Table 3** shows that GIS always outperforms denseCRF for the DSS model, except for MAE. However, the difference in MAE can be explained by the fact that denseCRF enhances the contrast.

Since the only differences between DSS-GIS and DSS are the added GIS layer from **Fig. 2**, the testing results show that it is fully justified to use the GIS layer instead of denseCRF since most improvements are due to it. However, using both together usually yields the best results.

**Table 2** Comparison between the percentage testing results of HED and our HED-GIS. The * means that a denseCRF layer is added at the output. The best value in each category is highlighted in bold (ignoring GIS*). The values are in percentages.

| Method | $AUC$ | $F_m$ | $MAE$ | $\overline{PR}$ | Dataset |
|---|---|---|---|---|---|
| HED-GIS* | +1.5 | +3.6 | -7.5 | +3.2 | ECSSD |
| HED* | +0.1 | +1.2 | -4.2 | +0.5 | |
| HED-GIS | **+1.5** | **+3.1** | **-6.0** | **+2.9** | |
| HED | 96.2 | 85.2 | 16.4 | 88.9 | |
| HED-GIS* | +3.5 | +8.7 | -9.0 | +8.7 | DUT-OMRON |
| HED* | +0.2 | +2.0 | -3.7 | +1.0 | |
| HED-GIS | **+3.5** | **+7.6** | **-7.3** | **+8.3** | |
| HED | 90.7 | 66.6 | 17.8 | 66.2 | |
| HED-GIS* | +1.3 | +3.0 | -7.2 | +2.6 | PASCAL-S |
| HED* | +0.1 | +1.1 | -4.1 | +0.5 | |
| HED-GIS | **+1.3** | **+2.4** | **-5.3** | **+2.5** | |
| HED | 92.4 | 77.1 | 20.5 | 79.7 | |

**Table 3** Comparison between the testing results of DSS and our DSS-GIS. The * means that a denseCRF layer is added at the output. The best value in each category is highlighted in bold (ignoring GIS*). The values are in percentages.

| Method | $AUC$ | $F_m$ | $MAE$ | $\overline{PR}$ | Dataset |
|---|---|---|---|---|---|
| DSS-GIS* | +0.1 | +0.7 | -0.7 | +0.6 | ECSSD |
| DSS* | -0.1 | +0.3 | **-0.6** | +0.0 | |
| DSS-GIS | **+0.3** | **+0.3** | -0.1 | **+0.6** | |
| DSS | 98.3 | 91.5 | 6.1 | 95.5 | |
| DSS-GIS* | +0.3 | +1.8 | -0.8 | +2.1 | DUT-OMRON |
| DSS* | -3.3 | +0.3 | **-1.0** | -2.6 | |
| DSS-GIS | **+0.5** | **+1.6** | -0.2 | **+2.0** | |
| DSS | 95.8 | 76.6 | 8.0 | 78.8 | |
| DSS-GIS* | +0.4 | +0.4 | -0.6 | +0.6 | PASCAL-S |
| DSS* | -0.2 | **+0.3** | **-0.7** | 0.0 | |
| DSS-GIS | **+0.6** | **+0.3** | 0.2 | **+0.6** | |
| DSS | 94.8 | 83.3 | 11.0 | 86.6 | |

## 5.3 Literature benchmarking

In this section, we aim at demonstrating that the proposed DSS-GIS algorithm outperforms other state-of-the-art (SoA) algorithms. The section first covers the comparison with algorithms trained on MSRA-B or MSRA10K [34], then it follows with algorithms trained on the more recent DUTS [38, 39] dataset. Then, it explores more complex problems such as saliency detection in very noisy images or low lighting images.

### 5.3.1 Training on MSRA10K

In **Table 4**, we show that our DSS-GIS algorithm outperforms all the other tested methods in terms of $F_m$ and $AUC$, with the exception of the $F_m$ on the PASCAL-S dataset, where it lacks by 0.1%. The improvements are mostly notable on the DUT-OMRON dataset since it is the most difficult one, with the most complex backgrounds and highest number of images. We also note that HED performs badly compared to other algorithms, while HED-GIS is very close to the high performance DCL method. The training of all algorithms is done on either the MSRA-B or MSRA10K dataset.

**Table 4** Comparison of the proposed DSS-GIS and HED-GIS approaches (grey rows) with other saliency algorithms proposed in the literature. The best result of each column is highlighted in bold. The values are in percentages. All methods are trained on either the MSRA-B or MSRA10K dataset.

| Dataset | ECSSD | | DUT-OMRON | | PASCAL-S | |
|---|---|---|---|---|---|---|
| Method | $F_m$ | $AUC$ | $F_m$ | $AUC$ | $F_m$ | $AUC$ |
| DSR [8] | 73.5 | 91.6 | 62.7 | 89.9 | 65.3 | 86.5 |
| RBD [11] | 71.6 | 89.6 | 62.9 | 89.2 | 65.9 | 85.8 |
| DRFI [13] | 78.5 | 94.5 | 66.5 | 93.2 | 70.0 | 89.9 |
| MDF [14] | 83.2 | 94.7 | 69.4 | 91.9 | 76.8 | 89.7 |
| MCDL [43] | 83.7 | 95.3 | 70.1 | 93.3 | 74.4 | 90.7 |
| UCF [44] | 90.3 | 98.2 | 73.0 | 94.6 | 82.3 | 95.0 |
| DCL [5] | 90.1 | 97.1 | 75.6 | 93.4 | 81.5 | 94.5 |
| Amulet [45] | 91.5 | 98.2 | 74.3 | 95.0 | **83.7** | 95.0 |
| HED [2] | 85.2 | 96.2 | 66.6 | 90.7 | 77.1 | 92.4 |
| HED-GIS | 88.3 | 97.7 | 74.1 | 94.3 | 79.5 | 93.7 |
| DSS [4] | 91.5 | 98.3 | 76.6 | 95.8 | 83.3 | 94.8 |
| DSS-GIS | **91.9** | **98.6** | **78.2** | **96.4** | 83.6 | **95.4** |

We also observed on **Fig. 11** that the precision/recall curves and the true-positive/false-positive curves of the DSS-GIS consistently outperforms the other methods. Again, HED-GIS outperforms HED by a high margin. The MCDL and UCF methods are not presented to improve the readability of the plot.

### 5.3.2 Training on the DUTS training set

Using the same parameter as the MSRA10K training, we trained the DSS model and DSS-GIS model on the DUTS training set, with the results reported in **Table 5** and validation curve in **Fig. 8**. As we can observe, the DSS performs much worse with the DUTS training set than with MSRA10K since it is very sensitive to changes of training conditions and training parameters as discussed previously in section 4.2. Hence its performance decreases since the DSS parameters were optimized for MSRA10K [42], and a new set of parameters need to be chosen.

This is not the case for DSS-GIS, which improved it's results from **Table 4**. Again, this is due to the ability of the GIS layer to make the network more robust to the training parameters, thanks to the improved regional representation via edge extrapolation. Hence, DSS-GIS strongly outperforms DSS with an average increase of 7.4% on $F_m$ and 3.4% on $AUC$ when using the DUTS training set without adapting the training parameters.

In **Table 5**, we observe that DSS-GIS outperforms AFNet on all $AUC$ measures with improvements of 1.0% to 2.4% and absolute values above 95.9% on all test sets. For the $F_m$, DSS-GIS outperforms AFNet of 1.0% and 1.2% on the hardest and biggest test sets (DUTS and DUT-OMRON have more than 5k images each), but lacks by 0.1% and 0.3% on the datasets with less than 1k images.

It is to note that DSS (on which DSS-GIS is based) is an older architecture based on Vgg16 [42], while AFNet is based on the more recent ResNet [40]. Hence, it demonstrates again that the simple addition of a GIS layer is more powerful than complex architectural changes. Future work could evaluate the improvements of the GIS layer on AFNet or other newer architectures.

**Table 5** Comparison of the proposed DSS-GIS approaches (grey rows) with AFNet, a recent saliency algorithms proposed in the literature trained on the DUTS training set. The best result of each column is highlighted in bold. The values are in percentages.

| Dataset | ECSSD | | DUT-OMRON | | PASCAL-S | | DUTS (test) | |
|---|---|---|---|---|---|---|---|---|
| Method | $F_m$ | $AUC$ | $F_m$ | $AUC$ | $F_m$ | $AUC$ | $F_m$ | $AUC$ |
| AFNet [40] | **93.5** | 98.0 | 78.4 | 93.5 | **86.6** | 94.6 | 85.7 | 96.5 |
| DSS [4] | 87.2 | 96.5 | 68.1 | 91.1 | 79.5 | 93.3 | 79.2 | 95.7 |
| DSS-GIS | 93.4 | **99.0** | **79.6** | **95.9** | 86.3 | **96.2** | **86.7** | **98.1** |

The the precision/recall curves and the true-positive/false-positive curves are also presented as the dotted lines in **Fig. 11**. We observe that DSS-GIS consistently outperforms AFNet in all true-positive/false-positive plots. For the precision/recall, DSS-GIS outperforms AFNet on all plots, except in the 75%-95% recall region of PASCAL-S were it is slightly below.

The validation curves of our model training are observed in **Fig. 8**, where our proposed DSS-GIS strongly outperforms DSS, with 3.7% higher $F_m$, 2.7 times lower loss and faster convergence. Further, we observe less noise on the DSS-GIS $F_m$ curve, implying that the model had more ease learning with the proposed GIS layer than without. The DSS-GIS reaches an $F_m$ of 93.7% in 100 iterations, which is the maximum reached by DSS after 29,900 iterations. The training parameters are exactly the same as for MSRA10K, and are described in section 2.2.3, which show that the GIS layer allowed the network to adapt to new training conditions.

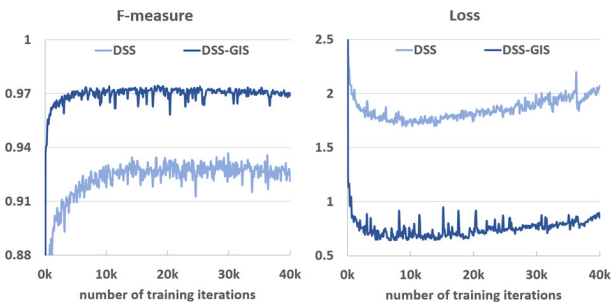

**Fig. 8** Comparison of the validation curves of our DSS-GIS against DSS on the DUTS training set with the same parameters as the MSRA10K training. The validation performance is computed every 100 iterations on the full validation set.

### 5.4 Resistance to noise and low-light

In the current section, we show that the proposed GIS layer allows the network to perform significantly better on the tests set when a high amount of noise is added, or when the brightness is significantly reduced.

To demonstrate it, we modify all images from the 3 testing sets by adding a 30% salt-and-pepper noise. We show in **Table 6** that the GIS layer significantly improves the $F_m$ and $AUC$ metrics. Furthermore, **Fig. 5** allows to observe this major difference, with the GIS layer allowing the model to find objects that were almost invisible to the standard method.

On standard images, the GIS layer only improved the ECSSD $F_m$ by 0.4%. However, on the noisy images, the improvement is 4.5%. On average for the DSS model, the GIS improves the $F_m$ by 3.9% and the $AUC$ by 3.0%. For the HED model, the average improvement is 8.1% on the $F_m$ and 10.7% on the $AUC$.

Additionally, we can observe in **Fig. 10** how the F-measure of proposed DSS-GIS is more stable than the DSS method for different levels of noise. The stronger the noise, the greater the margin between DSS-GIS and DSS.

The margin of improvement of the proposed GIS layer in a noisy setting is highly significant. This shows again the better generalizability of the saliency models since GIS allows the network to focus on the general features instead of very local noise and texture.

We believe that the major improvement is due to textures being more affected by the noise then edges, which plays in favor of the models implementing the GIS layer.

**Table 6** Comparison of the DSS and HED approaches with and without the proposed GIS layer when a 30% salt-and-pepper noise is added to the test set. The best result of each column is highlighted in bold. The values are in percentages. The training set is MSRA10K.

| Dataset | ECSSD | | DUT-OMRON | | PASCAL-S | |
|---|---|---|---|---|---|---|
| Method | $F_m$ | $AUC$ | $F_m$ | $AUC$ | $F_m$ | $AUC$ |
| HED [2] | 40.1 | 62.3 | 27.3 | 57.4 | 43.2 | 66.3 |
| HED-GIS | 50.6 | 73.0 | 39.0 | 71.6 | 45.3 | 73.4 |
| DSS [4] | 67.8 | 85.8 | 54.2 | 84.2 | 65.0 | 84.3 |
| DSS-GIS | **72.3** | **89.6** | **59.4** | **88.2** | **67.0** | **85.6** |

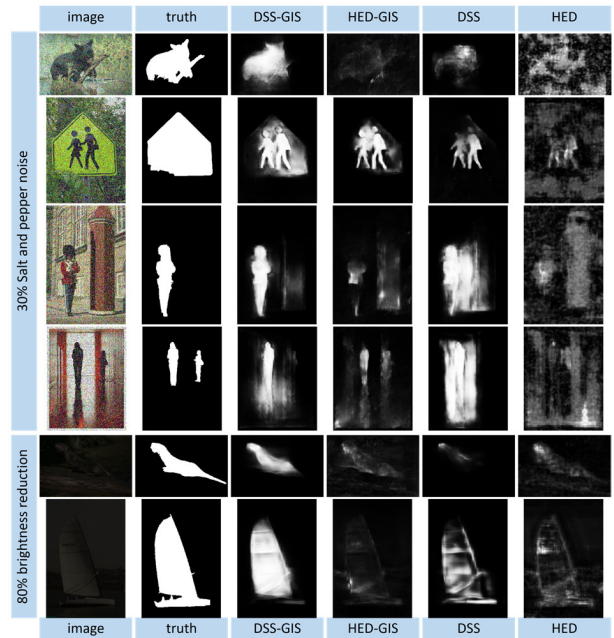

**Fig. 9** Test results comparison between the DSS and HED methods with and without the proposed GIS layer, when the test set is modified with noise or reduced brightness. The training set is MSRA10K.

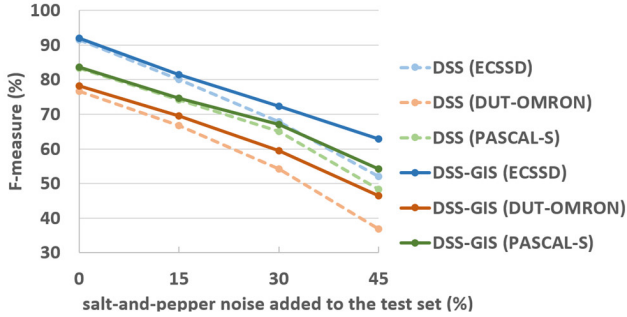

**Fig. 10** Performance impact of adding noise to the testing set for the DSS and the proposed DSS-GIS methods, trained on MSRA10K.

Additionally to the added robustness to noise, the proposed GIS layer allows the network to be more robust to other environmental changes, such as reduction in brightness. This is demonstrated in **Table 7** where DSS-GIS consistently outperforms DSS when the brightness is reduced by 80%, thus simulating an image taken at low light. Examples of such images are provided in **Fig. 9**. This improvement is due to the ability of the GIS-based networks to operate in the gradient-domain and to extrapolate edge information.

**Table 7** Comparison of the DSS and HED approaches with and without the proposed GIS layer when a the brightness is reduced by 80% to simulate low-light pictures. The best result of each column is highlighted in bold. The values are in percentages.

| Dataset | ECSSD | | DUT-OMRON | | PASCAL-S | |
|---|---|---|---|---|---|---|
| Method | $F_m$ | AUC | $F_m$ | AUC | $F_m$ | AUC |
| HED [2] | 68.2 | 87.4 | 53.2 | 83.6 | 60.9 | 84.2 |
| HED-GIS | 74.5 | 91.1 | 60.7 | 88.3 | 60.5 | 86.1 |
| DSS [4] | 83.0 | 93.3 | 69.8 | 91.1 | 74.4 | 88.2 |
| DSS-GIS | **84.5** | **95.8** | **72.6** | **94.4** | **76.9** | **91.4** |

### 5.5 Computation time

When using an image of the ECSSD dataset, the computation time for the DSS model is around 0.08s and the DSS-GIS is around 0.09s. Therefore, we see that the GIS layer has a low effect in terms of computation cost. Moreover, it was shown in **Fig. 7** that it improves the training repeatability and in **Table 3** and **Fig. 11** that it improves the F-measure on the testing results.

Other methods such as MDF [14] and DCL [5] require around 1s of computing due to the pre-segmentation, which is 10 times longer than the proposed DSS-GIS. Furthermore, the denseCRF layer takes around 0.4s to compute, which is 40 times longer than the added GIS layer. Hence, we suggest completely removing the denseCRF since it slows down the computation and leads to poorer performances than the GIS layer.

### 5.6 Future improvements

With the new GIS layer added at the end of the DSS network, the testing results are improved but by a moderate margin. One of the fundamental next steps is to take the same GIS layer, or other GFC-based layers, and to implement it deeper within the network, such as before the side layers or inside the VGGnet-16. Furthermore, the GIS layer should be tested for more applications such as classification, segmentation, and edge detection.

In fact, the GIS layers can be added to any other fully convolutional saliency architecture, not only the HED and DSS architecture as done in the current paper. Therefore, it adds good longevity to the GIS layer developed in this paper since newer architectures are also expected to benefit from the additional layer.

Finally, one of the most important contributions was showing that it is possible to add a Green's function convolution to a convolutional network to improve the results. This is surprising since CNN's usually have thousands of different and optimized convolutional kernels [25]. However, our work showed that a carefully engineered $V_{mono}$ convolutional kernel can still contribute to improving the results. This is because $V_{mono}$ adds a long-distance interaction between the pixel in the images, meaning that the receptive field is the whole image space. Also, since CNN are better at detecting edges than regions, integrating them into smooth and continuous regions naturally leads to better results.

For future work, we recommend using the same GIS network for segmentation purposes and for generative adversarial networks (GAN). In fact, we believe that the GIS layer would allow the GAN to generate image features in the gradient domain and the image domain at the same time. Since the GIS layer reduces noise sensitivity and gives the network an unlimited receptive field, we strongly believe that it can help generate better images. Such gradient-domain image drawing is already adopted by numerous software to allow drawing smoother images [20, 23, 27, 28].

### 5.7 Advantages of the GIS layer

The advantages of the GIS method compared to standard saliency improvement methods are summarized below.

**Simple and seamless integration** inside a CNN, by adding the GIS layer near the latest layer of any convolutional network, without architectural changes or increase in the number of parameters. We implemented GIS in both

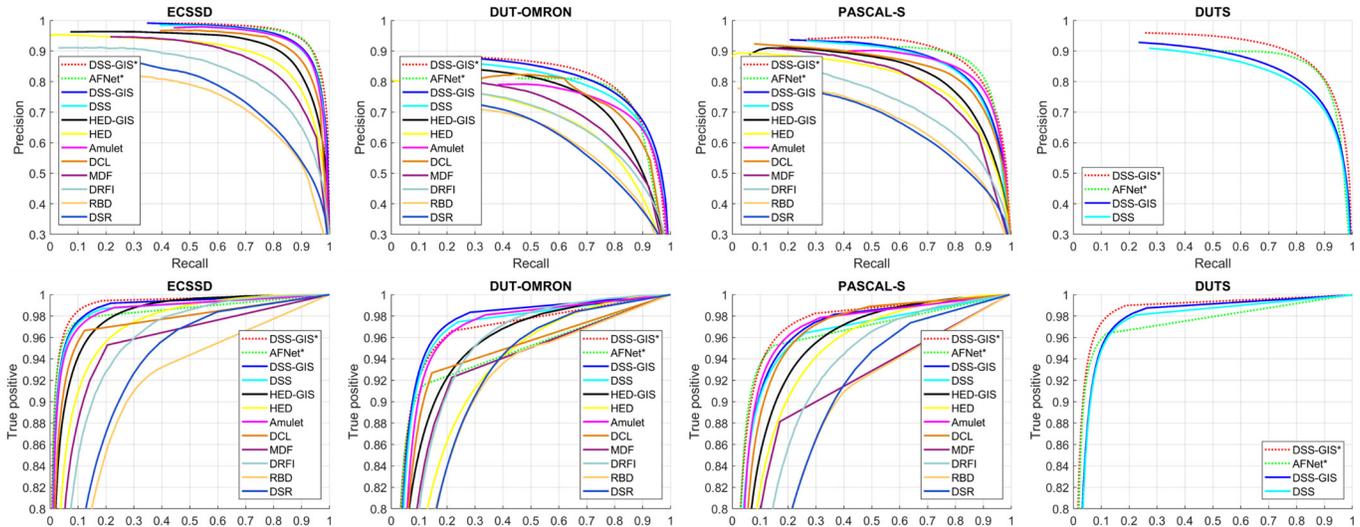

**Fig. 11** Precision-Recall curves (top row) and true-positive/false-positive curves (bottom row) for the 4 test datasets. The plain lines represent models trained on the MSRA10K dataset. The * and the dotted lines represent models trained on the DUTS training set.

Tensorflow [46] and Pytorch [47] by using the built-in FFT functions.

**Fast to compute**, with only 10ms increase in the computation time using the FFT with the GTX-1080Ti video card. This increase is negligible since most networks take between 70ms to 1200ms to compute the saliency map [3, 5, 14, 42, 45].

**Improves the network regional learning capacity** by allowing it to learn to fill edges into regions, without additional parameters or architectural changes. The GIS layer showed to improve the training repeatability, and to improve the testing results by a significant margin.

**Robust to noise and lighthing changes**, since GIS allows the network to identify salient images in settings where texture information is difficult to discern, but edges are easily identifiable.

We thus believe that the GIS brings novely to the field of saliency detection since, to the best of our knowledge, it is the first method to directly improve the networks learning capacity and generalization without architectural changes.

## 6 Conclusion

Our objective was to show that saliency convolutional networks can be improved by using a Green's function convolution (GFC) based layer to extrapolate edges features into salient regions. To answer this objective, we developed the gradient integration and sum (GIS) layer. We showed that using a GIS layer, inside both HED and DSS neural networks, improves the stability and repeatability of the training and enhances the performance of the model on the test set, with only 10 ms of added computation time. The GIS layer is fast to compute and does not require any weight or learned parameter. Moreover, the GIS layer reduces the training time, the overfitting, and the dependence to hyperparameters. GIS also makes the model significantly more resistant to noise and to different lighting conditions. The performance was generally better than other saliency enhancement methods such as denseCRF, 40 times faster to compute and directly integrated inside the network. Hence, our DSS-GIS network outperformed the tested state-of-the-art algorithms on most tested metrics such as the F-measure, the $AUC$ and the $MAE$. The GIS layer proved to improve the saliency by a higher margin than architectural improvement, since HED-GIS outperformed MDF and DSS-GIS outperformed AFNet.

The GIS method is novel since it increases the network regional learning capacity by allowing it to produce regions by filling edges. This contrasts with most methods which aim at post-processing the saliency map, or at changing the network architecture. A limitation of the current method is that it can only be used in the latest layers of a fully convolutional network for saliency purposes. Future work should experiment with implementing the GIS layer or other GFC-based layers deeper inside the network to try to further improve the results.